\documentclass[10pt,twocolumn,letterpaper]{article}

\usepackage[pagebackref=true,breaklinks=true,letterpaper=true,colorlinks,bookmarks=false]{hyperref}
\pdfoutput=1

\usepackage{cvpr}
\usepackage{times}
\usepackage{epsfig}
\usepackage{graphicx}
\usepackage{amsmath}
\usepackage{amssymb}

\usepackage{mathtools}

\usepackage{booktabs} 

\usepackage{authblk}

\cvprfinalcopy 

\def\cvprPaperID{2284} 

\ifcvprfinal\fi
\begin{document}

\title{Hardness-Aware Deep Metric Learning}

\author[1,2,3]{Wenzhao Zheng}
\author[1]{Zhaodong Chen}
\author[1,2,3]{Jiwen Lu\thanks{Corresponding author}}
\author[1,2,3]{Jie Zhou}
\affil[1]{Department of Automation, Tsinghua University, China }
\affil[2]{State Key Lab of Intelligent Technologies and Systems, China}
\affil[3]{Beijing National Research Center for Information Science and Technology, China}
\affil[ ]{\tt\small zhengwz18@mails.tsinghua.edu.cn; chenzd15@mails.tsinghua.edu.cn; lujiwen@tsinghua.edu.cn; jzhou@tsinghua.edu.cn}

\maketitle

\begin{abstract}
This paper presents a hardness-aware deep metric learning (HDML) framework. Most previous deep metric learning methods employ the hard negative mining strategy to alleviate the lack of informative samples for training. However, this mining strategy only utilizes a subset of training data, which may not be enough to characterize the global geometry of the embedding space comprehensively. To address this problem, we perform linear interpolation on embeddings to adaptively manipulate their hard levels and generate corresponding label-preserving synthetics for recycled training, so that information buried in all samples can be fully exploited and the metric is always challenged with proper difficulty. Our method achieves very competitive performance on the widely used CUB-200-2011, Cars196, and Stanford Online Products datasets.
\footnote{Code: \url{https://github.com/wzzheng/HDML}}
\end{abstract}

\section{Introduction}
Deep metric learning methods aim to learn effective metrics to measure the similarities between data points accurately and robustly. They take advantage of deep neural networks~\cite{krizhevsky2012imagenet, simonyan2014very, szegedy2015going, he2016deep} to construct a mapping from the data space to the embedding space so that the Euclidean distance in the embedding space can reflect the actual semantic distance between data points, i.e., a relatively large distance between inter-class samples and a relatively small distance between intra-class samples. Recently a variety of deep metric learning methods have been proposed and have demonstrated strong effectiveness in various tasks, such as image retrieval~\cite{song2016deep, opitz2017bier, law2017deep,duan2018deep}, person re-identification~\cite{shi2016embedding, wang2016joint, zhou2017efficient, chen2017beyond}, and geo-localization~\cite{vo2016localizing, kim2017learned, vo2017revisiting}. 

\begin{figure}[tb]
\centering
\includegraphics[width=0.475\textwidth]{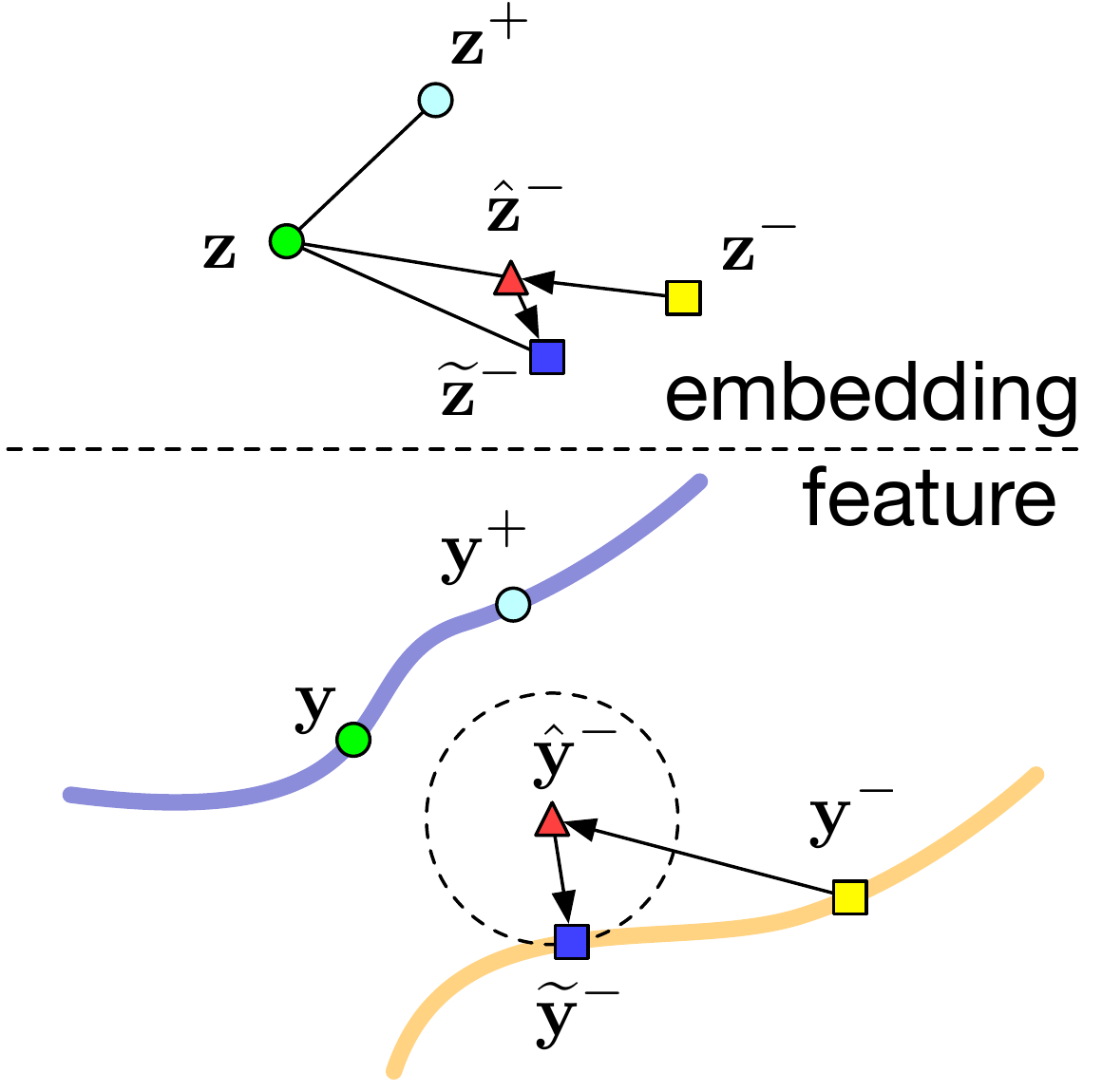}
\caption{Illustration of our proposed hardness-aware feature synthesis. A curve in the feature space represents a manifold near which samples belong to one specific class concentrate. Points with the same color in the feature space and embedding space represent the same sample and points of the same shape denote that they belong to the same class. The proposed hardness-aware augmentation first modifies a sample $\mathbf{y}^-$ to $\hat{\mathbf{y}}^-$. Then a label-and-hardness-preserving generator projects it to $\widetilde{\mathbf{y}}^-$ which is the closest point to $\hat{\mathbf{y}}^-$ on the manifold. The hardness of synthetic negative $\widetilde{\mathbf{y}}^-$ can be controlled adaptively and does not change the original label so that the synthetic hardness-aware tuple can be favorably exploited for effective training. (Best viewed in color.)}
\label{fig: manifold}
\end{figure}

The overall training of a deep metric learning model can be considered as using a loss weighted by the selected samples, which makes the sampling strategy a critical component. A primary issue concerning the sampling strategy is the lack of informative samples for training. A large fraction of samples may satisfy the constraints imposed by the loss function and provide 
no supervision information for the training model. This motivates many deep metric learning methods to develop efficient hard negative mining strategies~\cite{schroff2015facenet, huang2016local, yuan2017hard, harwood2017smart} for sampling. These strategies typically under-sample the training set for hard informative samples which produce gradients with large magnitude.
However, the hard negative mining strategy only selects among a subset of samples, which may not be enough to characterize the global geometry of the embedding space accurately. In other words, some data points are sampled repeatedly while others may never have the possibility to be sampled, resulting in an embedding space over-fitting near the over-sampled data points and at the same time under-fitting near the under-sampled data points.

In this paper, we propose a hardness-aware deep metric learning (HDML) framework as a solution. We sample all data points in the training set uniformly while making the best of the information contained in each point. Instead of only using the original samples for training, we propose to synthesize hardness-aware samples as complements to the original ones. In addition, we control the hard levels of the synthetic samples according to the training status of the model, so that the better-trained model is challenged with harder synthetics. We employ an adaptive linear interpolation method to effectively manipulate the hard levels of the embeddings. Having obtained the augmented embeddings, we utilize a simultaneously trained generator to map them back to the feature space while preserving the label and augmented hardness. These synthetics contain more information than original ones and can be used as complements for recycled training, as shown in Figure \ref{fig: manifold}.
We provide an ablation study to demonstrate the effectiveness of each module of HDML. Extensive experiments on the widely-used CUB-200-2011~\cite{wah2011caltech}, Cars196~\cite{krause20133d}, and Stanford Online Products~\cite{song2016deep} datasets illustrate that our proposed HDML framework can improve the performance of existing deep metric learning models in both image clustering and retrieval tasks.

\section{Related Work}

\textbf{Metric Learning:} Conventional metric learning methods usually employ the Mahalanobis distance~\cite{globerson2006metric, davis2007information, weinberger2009distance} or kernel-based metric~\cite{frome2007learning} to characterize the linear and non-linear intrinsic correlations among data points. Contrastive loss~\cite{hadsell2006dimensionality, hu2014discriminative} and triplet loss~\cite{wang2014learning, schroff2015facenet, cheng2016person} are two conventional measures which are widely used in most existing metric learning methods. The contrastive loss is designed to separate samples of different classes with a fixed margin and pull closer samples of the same category as near as possible.
The triplet loss is more flexible since it only requires a certain ranking within triplets. Furthermore, there are also some works to explore the structure of quadruplets~\cite{law2013quadruplet, huang2016local, chen2017beyond}.

The losses used in recently proposed deep metric learning methods~\cite{song2016deep, sohn2016improved, ustinova2016learning, song2017deep, wang2017deep, yu2018correcting} take into consideration of higher order relationships or global information and therefore achieve better performance. For example, Song~\emph{et al.}~\cite{song2016deep} proposed a lifted structured loss function to consider all the positive and negative pairs within a batch. Wang~\emph{et al.}~\cite{wang2017deep} improved the conventional triplet loss by exploiting a third-order geometry relationship. These meticulously designed losses showed great power in various tasks, yet a more advanced sampling framework~\cite{wu2017sampling, movshovitz2017no, ge2018deep,lin2018deep} can still boost their performance. For example, Wu~\emph{et al.}~\cite{wu2017sampling} presented a distance-weighted sampling method to select samples based on their relative distances. Another trend is to incorporate ensemble technique in deep metric learning~\cite{opitz2017bier,kim2018attention,xuan2018deep}, which integrates several diverse embeddings to constitute a more informative representation.

\textbf{Hard Negative Mining:} Hard negative mining has been employed in many machine learning tasks to enhance the training efficiency and boost performance, like supervised learning~\cite{schroff2015facenet, huang2016local, yuan2017hard, harwood2017smart, yu2018hard}, exemplar based learning~\cite{malisiewicz2011ensemble} and unsupervised learning~\cite{wang2015unsupervised, bautista2017deep}.
This strategy aims at progressively selecting false positive samples that will benefit training the most. It is widely used in deep metric learning methods because of the vast number of tuples that can be formed for training. 
For example, Schroff~\emph{et al.}~\cite{schroff2015facenet} proposed to sample “semi-hard” triplets within a batch, which avoids using too confusing triplets that may result from noisy data. Harwood~\emph{et al.}~\cite{harwood2017smart} presented a smart mining procedure utilizing approximate nearest neighbor search methods to adaptively select more challenging samples for training. The advantage of \cite{yuan2017hard} and \cite{harwood2017smart} lies in the selection of samples with suitably hard level with the model. However, they can not control the hard level accurately and do not exploit the information contained in the easy samples. 

Recently proposed methods~\cite{duan2018deep,zhao2018adversarial} begin to consider generating potential hard samples to fully train the model. However, there are several drawbacks of the current methods. Firstly, the hard levels of the generated samples cannot be controlled. Secondly, they all require an adversarial manner to train the generator, rendering the model hard to be learned end-to-end and the training process very unstable. Differently, the proposed HDML framework can generate synthetic hardness-aware label-preserving samples with adequate information and adaptive hard levels, further boosting the performance of current deep metric learning models.

\section{Proposed Approach}

In this section, we first formulate the problem of deep metric learning and then present the basic idea of the proposed HDML framework. At last, we elaborate on the approach of deep metric learning under this framework. 

\begin{figure}[tb]
\centering
\includegraphics[width=0.475\textwidth]{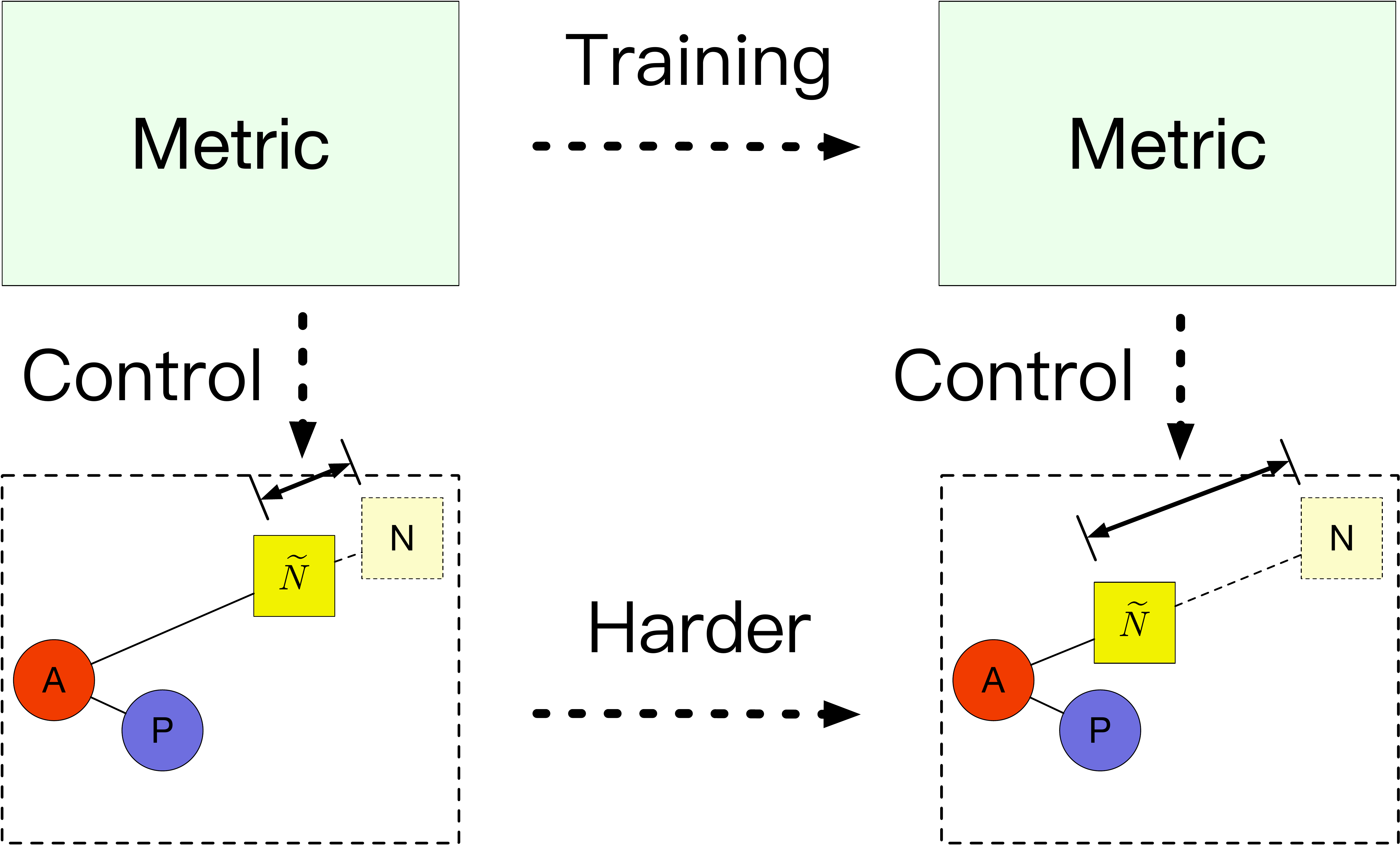}
\caption{Illustration of the proposed hardness-aware augmentation. Points with the same shape are from the same class. We performs linear interpolation on the negative pair in the embedding space to obtain a harder tuple, where the hard level is controlled by the training status of the model. As the training proceeds, harder and harder tuples are generated to train the metric more efficiently. 
(Best viewed in color.)
} 
\label{model1}
\end{figure}

\subsection{Problem Formulation}
Let $\pmb{\mathcal{X}}$ denote the data space where we sample a set of data points $\mathbf{X}=[\mathbf{x}_1, \mathbf{x}_2, \cdots , \mathbf{x}_N]$. Each point $\mathbf{x}_i$ has a label $l_i \in \{1, \cdots, C\}$ which constitutes the label set $\mathbf{L}=[l_1, l_2, \cdots , l_N]$. Let $f : \pmb{\mathcal{X}} \xrightarrow{f} \pmb{\mathcal{Y}}$ be a mapping from the data space to a feature space, where the extracted feature $\mathbf{y}_i$ has semantic characteristics of its corresponding data point $\mathbf{x}_i$. The objective of metric learning is to learn a distance metric in the feature space so that it can reflect the actual semantic distance. The distance metric can be defined as:
\begin{eqnarray}\label{distance metric}
D(\mathbf{x}_i,\mathbf{x}_j) = m(\theta_m; \mathbf{y}_i,\mathbf{y}_j) = m(\theta_m; f(\mathbf{x}_i),f(\mathbf{x}_j)),
\end{eqnarray}
where $m$ is a consistently positive symmetric function and $\theta_m$ is the corresponding parameters.

\begin{figure*}[tb]
\centering
\includegraphics[width=1\textwidth]{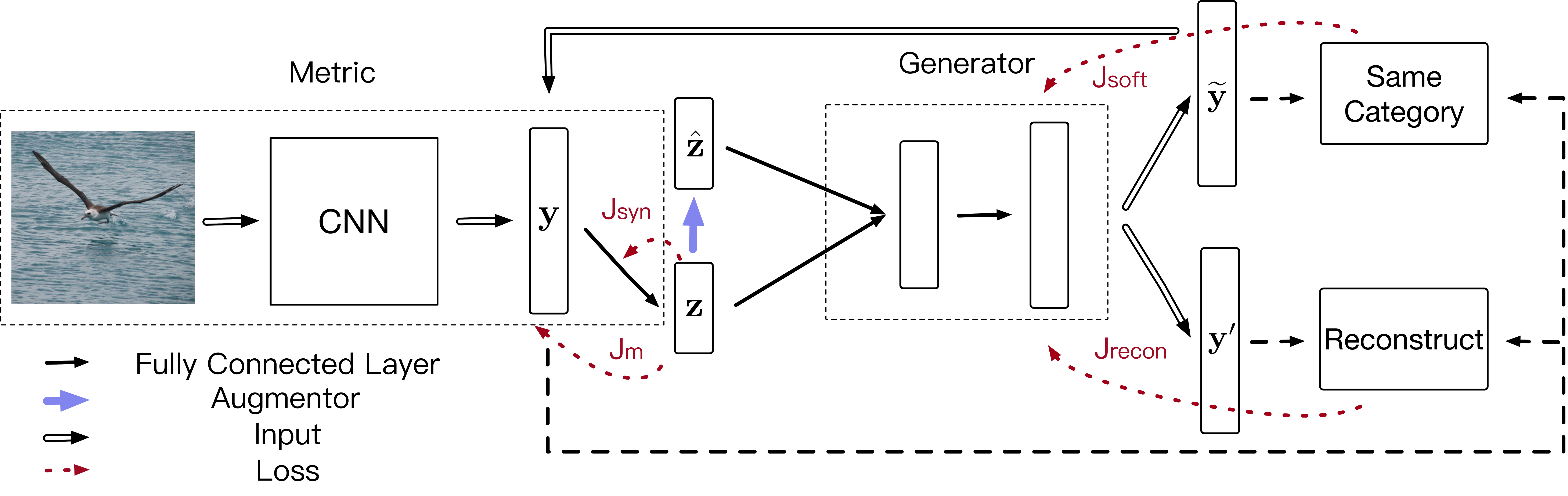}
\caption{The overall network architecture of the our HDML framework. The red dashed arrow points from the part that the loss is computed on, and to the module that the loss directly supervises. The metric model is a CNN network followed by a fully connected layer. The augmentor is a linear manipulation of the input and the generator is composed of two fully connected layers with increasing dimensions. Part of the metric and the following generator form a similar structure to the well-known autoencoder. (Best viewed in color.)} 
\label{model2}
\end{figure*}

Deep learning methods usually extract features using a deep neural network. A standard procedure is to first project the features into an embedding space (or metric space) $\pmb{\mathcal{Z}}$ with a mapping $g : \pmb{\mathcal{Y}} \xrightarrow{g} \pmb{\mathcal{Z}}$, where the distance metric is then a simple Euclidean distance. 
Since the projection can be incorporated into the deep network, we can directly learn a mapping $h = g \circ f: \pmb{\mathcal{X}} \xrightarrow{h} \pmb{\mathcal{Z}}$ from the data space to the embedding space, so that the whole model can be trained end-to-end without explicit feature extraction. In this case, the distance metric is defined as:
\begin{eqnarray}\label{deep learning distance metric}
D(\mathbf{x}_i,\mathbf{x}_j) = d(\mathbf{z}_i,\mathbf{z}_j) = d(\theta_h; h(\mathbf{x}_i),h(\mathbf{x}_j)),
\end{eqnarray}
where $d$ indicates the Euclidean distance $d(\mathbf{z}_i, \mathbf{z}_j) = ||\mathbf{z}_i -\mathbf{z}_j||_2$, $\mathbf{z} = g(\mathbf{y}) = h(\mathbf{x})$ is the learned embedding, $\theta_f$, $\theta_g$ and $\theta_h$ are the parameters of mappings $f$, $g$ and $h$ respectively, and $\theta_h = \{\theta_f, \theta_g\}$.   

Metric learning models are usually trained based on tuples $\{\mathbf{T}_i\}$ composed of several samples with certain similarity relations. The network parameters are learned by minimizing a specific loss function:
\begin{eqnarray}\label{loss function}
\theta_h^* = \mathop{\arg\min}_{\theta_h} \ \  {J} (\theta_h; \{\mathbf{T}_i\}).
\end{eqnarray}

For example, the triplet loss~\cite{schroff2015facenet} samples triplets consisting of three examples, the anchor $\mathbf{x}$, the positive $\mathbf{x^+}$ with the same label with the anchor, and the negative $\mathbf{x^-}$ with a different label. The triplet loss forces the distance between the anchor and the negative to be larger than the distance between the anchor and the positive by a fixed margin.

Furthermore, the N-pair Loss~\cite{sohn2016improved} samples tuples with $N$ positive pairs of distinctive classes, and attempts to push away $N-1$ negatives altogether.

\subsection{Hardness-Aware Augmentation}
There may exist a great many tuples that can be used during training, yet the vast majority of them actually lack direct information and produce gradients that are approximately zero. To only select among the informative ones we limit ourselves to a small set of tuples. However, this small set may not be able to accurately characterize the global geometry of the embedding space, leading to a biased model. 

To address the above limitations, we propose an adaptive hardness-aware augmentation method, as shown in Figure \ref{model1}. We modify and construct the hardness-aware tuples in the embedding space, where manipulation of the distances among samples will directly alter the hard level of the tuple. A reduction in the distance between negative pairs will create a rise of the hard level and vice versa.

Given a set we can usually form more negative pairs than positive pairs, so for simplicity, we only manipulate the distances of negative pairs. For other samples in the tuple, we perform no transformation, i.e., $\hat{\mathbf{z}} = \mathbf{z}$. Still, our model can be easily extended to deal with positive pairs. 
Having obtained the embeddings of a negative pair (an anchor $\mathbf{z}$ and a negative $\mathbf{z^-}$), we construct an augmented harder negative sample $\mathbf{\hat{z}^-}$ by linear interpolation:
\begin{eqnarray}\label{equ: lambda_0_range}
\mathbf{\hat{z}^-} =\mathbf{z} + \lambda_0(\mathbf{z^-} - \mathbf{z}), \lambda_0 \in [0, 1].
\end{eqnarray}
However, an example too close to the anchor is very likely to share the label, thus no longer constitutes a negative pair. Therefore, it is more reasonable to set $\lambda_0 \in (\frac{d^+}{d(\mathbf{z}, \mathbf{z^-})}, 1]$, where $d^+$ is a reference distance that we use to determine the scale of manipulation (e.g., the distance between a positive pair or a fixed value), and $d(\mathbf{z}, \mathbf{z^-}) = ||\mathbf{z}^- -\mathbf{z}||_2$. To achieve this, we introduce a variable $\lambda \in (0,1]$ and set
\begin{eqnarray}\label{equ: lambda_0_range}
\lambda_0 =
\left\{  
             \begin{array}{lr}  
                \lambda + (1-\lambda)\frac{d^+}{d(\mathbf{z}, \mathbf{z^-})} &, \ \ if\ \  d(\mathbf{z},\mathbf{z^-}) > d^+   \\  
              1   &, \ \ if \ \ d(\mathbf{z},\mathbf{z^-}) \le d^+.
             \end{array}  
\right. 
\end{eqnarray}
On condition that $d(\mathbf{z},\mathbf{z^-}) > d^+$, the augmented negative sample can be presented as:
\begin{equation}\label{equ: general_aug}
    \hat{\mathbf{z}}^- = \mathbf{z}+ [\lambda d(\mathbf{z}, \mathbf{z^-}) + (1-\lambda)d^+]\frac{\mathbf{z^-}-\mathbf{z}}{d(\mathbf{z}, \mathbf{z^-})}.
\end{equation}

Since the overall hardness of original tuples gradually decreases during training, it's reasonable to increase progressively the hardness of synthetic tuples for compensation. The hardness of a triplet increases when $\lambda$ gets smaller, so we can intuitively set $\lambda$ to $e^{-\frac{\alpha}{J_{avg}}}$, 
where $J_{avg}$ is the average metric loss over the last epoch, and $\alpha$ is the pulling factor used to balance the scale of $J_{avg}$. 
We exploit the average metric loss to control the hard level since it is a good indicator of the training process. The augmented negative is closer to the anchor if a smaller average loss, leading to harder tuples as training proceeds. 
The proposed hardness-aware negative augmentation can be represented as:
\begin{small}
\begin{eqnarray}\label{equ: hardness-aware extrapolation}
\hat{\mathbf{z}}^- =
\left\{  
             \begin{array}{lr}  
               \mathbf{z} + [ e^{-\frac{\alpha}{J_{avg}}} d(\mathbf{z},\mathbf{z}^-) + (1-e^{-\frac{\alpha}{J_{avg}}})d^+ ] \frac{\mathbf{z}^- - \mathbf{z}}{d(\mathbf{z},\mathbf{z}^-)} \\
               ~~~~~~~~~~~~~~~~~~~~~~~~~~~~~~~~~~~~~~~~~~~~~~~~~~~~if~ d(\mathbf{z},\mathbf{z^-}) > d^+   \\  
              \mathbf{z}^- ~~~~~~~~~~~~~~~~~~~~~~~~~~~~~~~~~~~~~~~~~~~~~~~if~ d(\mathbf{z},\mathbf{z^-}) \le d^+.
             \end{array}  
\right. 
\end{eqnarray}
\end{small}

The necessity of adaptive hardness-aware synthesis lies in two aspects. Firstly, in the early stages of training, the embedding space does not have an accurate semantic structure, so currently hard samples may not truly be informative or meaningful, and hard synthetics in this situation may be even inconsistent. Also, hard samples usually result in significant changes of the network parameters. Thus the use of meaningless ones can easily damage the embedding space structure, leading to a model that is trained in the wrong direction from the beginning. On the other hand, as the training proceeds, the model is more tolerant of hard samples, so harder and harder synthetics should be generated to keep the learning efficiency at a high level.

\subsection{Hardness-and-Label-Preserving Synthesis}
Having obtained the hardness-aware tuple in the embedding space, our objective is to map it back to the feature space so they can be exploited for training. However, this mapping is not trivial, since a negative sample constructed following \eqref{equ: hardness-aware extrapolation} may not necessarily benefit the training process: there is no guarantee that $\mathbf{\hat{z}}^-$ shares the same label with $\mathbf{z}^-$. To address this, we formulate this problem from a manifold perspective, and propose a hardness-and-label-preserving feature synthesis method.

As shown in Figure \ref{fig: manifold}, the two curves in the feature space represent two manifolds near which the original data points belong to class $l$ and $l^-$ concentrate respectively. Points with the same color in the feature and embedding space represent the same example. So below we do not distinguish operations acting on features and embeddings. $\mathbf{y}_n$ is a real data point of class $l_n$, and we first augment it to $\mathbf{\hat{y}^-}$ following \eqref{equ: hardness-aware extrapolation}. $\mathbf{\hat{y}^-}$ is more likely to be outside and further from the manifold compared with original data points since it is close to $\mathbf{y}$ that belongs to another category. Intuitively, the goal is to learn a generator that maps $\mathbf{\hat{y}^-}$, a data point away from the manifold (less likely belonging to class $l^-$), to a data point that lies near the manifold (more likely belonging to class $l^-$). Moreover, to best preserve the hardness, this mapped point should be close to $\mathbf{\hat{y}^-}$ as much as possible. These two conditions restrict the target point to $\mathbf{\widetilde{y}^-}$, which is the closest point to $\mathbf{\hat{y}^-}$ on the manifold.

We achieve this by learning a generator $i : \pmb{\mathcal{Z}} \xrightarrow{i} \pmb{\mathcal{Y}}$, which maps the augmented embeddings of a tuple back to the feature space for recycled training. Since a generator usually cannot perfectly map all the embeddings back to the feature space, the synthetic features must lie in the same space to provide meaningful information. Therefore, we map not only the synthetic negative sample but also the other unaltered samples in one tuple:
\begin{eqnarray}\label{transformed}
\mathbf{T}(\widetilde{\mathbf{y}})= i(\theta_i;  \mathbf{T}(\hat{\mathbf{z}})),
\end{eqnarray}
where $\mathbf{T}(\widetilde{\mathbf{y}})$ and $\mathbf{T}(\hat{\mathbf{z}})$ are tuples in the feature and embedding space respectively, and $\theta_i$ is the parameters of the generative mapping $i$.

We exploit an auto-encoder architecture to implement the mapping $g$ and mapping $i$.
The encoder $g$ takes as input a feature vector $\mathbf{y}$ which is extracted by CNN from the image, and first maps it to an embedding $\mathbf{z}$. In the embedding space, we modify $\mathbf{z}$ to $\mathbf{\hat{z}}$ using the hardness-aware augmentation described in the last subsection. The generator $i$ then maps the original embedding $\mathbf{z}$ and the augmented embedding $\mathbf{\hat{z}}$ to $\mathbf{y'}$ and $\mathbf{\widetilde{y}}$ respectively. 

In order to exploit the synthetic features $\mathbf{\widetilde{y}}$ for effective training, they should preserve the labels of the original samples as well as the augmented hardness. We formulate the objective of the generator as follows:
\begin{eqnarray}\label{equ: auto-encoder_loss}
J_{gen} &=& J_{recon} + \lambda J_{soft} \nonumber \\
&=& c(\mathbf{Y}, \mathbf{Y'}) + \lambda J_{soft} (\widetilde{\mathbf{Y}}, \mathbf{L})  \nonumber \\
&=& \sum_{\substack{\mathbf{y} \in \mathbf{Y} \\ \mathbf{y'} \in \mathbf{Y'}}} || \mathbf{y} - \mathbf{y'}||^2 + \lambda \sum_{\substack{\widetilde{\mathbf{y}} \in \widetilde{\mathbf{Y}} \\ l \in \mathbf{L} }} j_{soft} (\widetilde{\mathbf{y}}, l),
\end{eqnarray}
where $\lambda$ is a balance factor, $\mathbf{y'} = i(\theta_i; \mathbf{z})$ is the \emph{unaltered} synthetic feature, $\widetilde{\mathbf{y}}$ is the hardness-aware synthetic feature of origin $\mathbf{y}$ with label $l$, $\mathbf{Y'}$, $\mathbf{Y}$ and $\widetilde{\mathbf{Y}}$ are the corresponding feature distributions, $c(\mathbf{Y}, \mathbf{Y'})$ is the reconstruction cost between the two distributions, and $J_{soft}$ is the softmax loss function. Note that $J_{gen}$ is only used to train the decoder/generator and has no influence on the metric.

The overall objective function is composed of two parts: the reconstruction loss and the softmax loss. The synthetic negative should be as close to the augmented negative as possible so that it can constitute a tuple with hardness we require. Thus we utilize the reconstruction loss $J_{recon} =|| \mathbf{y} - \mathbf{y'}||^2_2$ to restrict the encoder \& decoder to map each point close to itself.
The softmax loss $J_{soft}$ ensures that the augmented synthetics do not change the original label. Directly penalizing the distance between $\widetilde{y}$ and $y$ can also achieve this, but is too strict to preserve the hardness. Alternatively, we simultaneously learn a fully connected layer with the softmax loss on $\mathbf{y}$, where the gradients only update the parameters in this layer. We employ the learned softmax layer to compute the softmax loss $j_{soft} (\widetilde{\mathbf{y}}, l)$ between the synthetic hardness-aware negative $\mathbf{\widetilde{y}}$ and the original label $l$.

\subsection{Hardness-Aware Deep Metric Learning}

We present the framework of the proposed method, which is mainly composed of three parts, a metric network to obtain the embeddings, a hardness-aware augmentor to perform augmentation of the hard level and a hardness-and-label-preserving generator network to generate the corresponding synthetics, as shown in Figure~\ref{model2}.

Having obtained the embeddings of a tuple, we first perform linear interpolation to modify the hard level, weighted by a factor indicating the current training status of the model. Then we utilize a simultaneously trained generator to generate synthetics for the augmented hardness-aware tuple, meanwhile ensuring the synthetics are realistic and maintain their original labels. Compared to conventional deep metric learning methods, we additionally utilize the hardness-aware synthetics to train the metric:
\begin{eqnarray}\label{loss function2}
\theta_h^* = \mathop{\arg\min}_{\theta_h} \ \  {J} (\theta_h; \{\mathbf{T}_i\} \cup \{\widetilde{\mathbf{T}}_i\}),
\end{eqnarray}
where $\widetilde{\mathbf{T}}_i$ is the synthetic hardness-aware tuple.

The proposed framework can be applied to a variety of deep metric learning methods to boost their performance. For a specific loss $J$ in metric learning, the objective function to train the metric is:
\begin{eqnarray}\label{transformed}
J_{metric} &=& e^{-\frac{\beta}{J_{gen}}} J_m + (1-e^{-\frac{\beta}{J_{gen}}}) J_{syn} \nonumber \\
&=& e^{-\frac{\beta}{J_{gen}}} J(\mathbf{T}) + (1-e^{-\frac{\beta}{J_{gen}}}) J(\widetilde{\mathbf{T}}),
\end{eqnarray}
where $\beta$ is a pre-defined parameter, $J_m = J(\mathbf{T})$ is the loss $J$ over original samples, $J_{syn} = J(\widetilde{\mathbf{T}})$ is the loss $J$ over synthetic samples, and $\widetilde{\mathbf{T}}$ denotes the synthetic tuple in the feature space. We use $e^{-\frac{\beta}{J_{gen}}}$ as the balance factor to assign smaller weights to synthetic features when $J_{gen}$ is high, since the generator is not fully trained and the synthetic features may not have realistic meanings.

$J_m$ aims to learn the embedding space so that inter-class distances are large and intra-class distances are small. $J_{syn}$ utilizes synthetic hardness-aware samples to train the metric more effectively. As the training proceeds, harder tuples are synthesized to keep the high efficiency of learning.

We demonstrate our framework on two losses with different tuple formations: triplet loss~\cite{schroff2015facenet} and N-pair loss~\cite{sohn2016improved}. 

For the triplet loss~\cite{schroff2015facenet}, we use the distance of the positive pair as the reference distance and generate the negative with our hardness-aware synthesis: 
\begin{eqnarray}\label{transformed}
J(\widetilde{\mathbf{T}}(\mathbf{x}, \mathbf{x}^+, \widetilde{\mathbf{x}}^-)) = [D(\mathbf{x},\mathbf{x}^+) - D(\mathbf{x},\widetilde{\mathbf{x}}^-) + m]_+,
\end{eqnarray}
where $[\cdot]_+ = max(\cdot, 0)$ and $m$ is the margin.

For the N-pair loss~\cite{sohn2016improved}, we also use the distance of the positive pair as the reference distance, but generate all the $N-1$ negatives for each anchor in an (N+1)-tuple: 
\begin{eqnarray}
&&J(\widetilde{\mathbf{T}}(\{\mathbf{x},\mathbf{x}^+, \widetilde{\mathbf{x}}^+\}_i)) \\
 &=& \frac{1}{N}\sum_{i=1}^N\log{(1+\sum_{j\neq i}\exp{(D(\mathbf{x}_i,\mathbf{x}_i^+)- D(\mathbf{x}_i,\widetilde{\mathbf{x}}_j^+))})}. \nonumber
\end{eqnarray}

The metric and the generator network are trained simultaneously, without any interruptions for auxiliary sampling processes as most hard negative mining methods do. The augmentor and generator are only used in the training stage, which introduces no additional workload to the resulting embedding computing.

\section{Experiments}
In this section, we conducted various experiments to evaluate the proposed HDML in both image clustering and retrieval tasks.
We performed an ablation study to analyze the effectiveness of each module. For the clustering task, we employed NMI and $\mathrm{F}_1$ as performance metrics. The normalized mutual information (NMI) is defined by the ratio of the mutual information of clusters and ground truth labels and the arithmetic mean of their entropy. $\mathrm{F}_1$ is the harmonic mean of precision and recall. See~\cite{song2016deep} for more details. For the retrieval task, we employed Recall@Ks as performance metrics. They are determined by the existence of at least one correct retrieved sample in the K nearest neighbors. 

\subsection{Datasets}
We evaluated our method under a zero-shot setting, where the training set and test set contain image classes with no intersection. We followed~\cite{song2016deep,song2017deep,duan2018deep} to perform the training/test set split.

\begin{itemize}

\item The CUB-200-2011 dataset~\cite{wah2011caltech} consists of 11,788 images of 200 bird species. We split the first 100 species (5,864 images) for training and the rest 100 species (5,924 images) for testing. 
\item The Cars196 dataset~\cite{krause20133d} consists of 16,185 images of 196 car makes and models. We split the first 98 models (8,054 images) for training and the rest 100 models (8,131 images) for testing. 
\item The Stanford Online Products dataset~\cite{song2016deep} consists of 120,053 images of 22,634 online products from eBay.com. We split the first 11,318 products (59,551 images) for training and the rest 11,316 products (60,502 images) for testing. 

\end {itemize}

\subsection{Experimental Settings}
We used the Tensorflow package throughout the experiments. For a fair comparison with previous works on deep metric learning, we used GoogLeNet~\cite{szegedy2015going} architecture as the CNN feature extractor (i.e., $f$) and added a fully connected layer as the embedding projector (i.e., $g$). We implemented the generator (i.e., $i$) with two fully connected layers of increasing output dimensions 512 and 1,024. We fixed the embedding size to 512 for all the three datasets. 
For training, we initialized the CNN with weights pre-trained on ImageNet ILSVRC dataset~\cite{russakovsky2015imagenet} and all other fully connected layers with random weights. We first resized the images to 256 by 256, then performed random cropping at 227 by 227 and horizontal random mirror for data augmentation. We tuned all the hyperparameters via 5-fold cross-validation
on the training set. We set the learning rate for CNNs to $10^{-4}$ and multiplied it by 10 for other fully connected layers. We set the batch size to 120 for the triplet loss and 128 for the N-pair loss. We fixed the balance factors $\beta$ and $\lambda$ to $10^4$ and $0.5$, and set $\alpha$ to 7 for the triplet loss and 90 for the N-pair loss.

\subsection{Results and Analysis}

\newcommand\figwidth{0.45}

\begin{figure}[tb]
\centering
\includegraphics[width=\figwidth\textwidth]{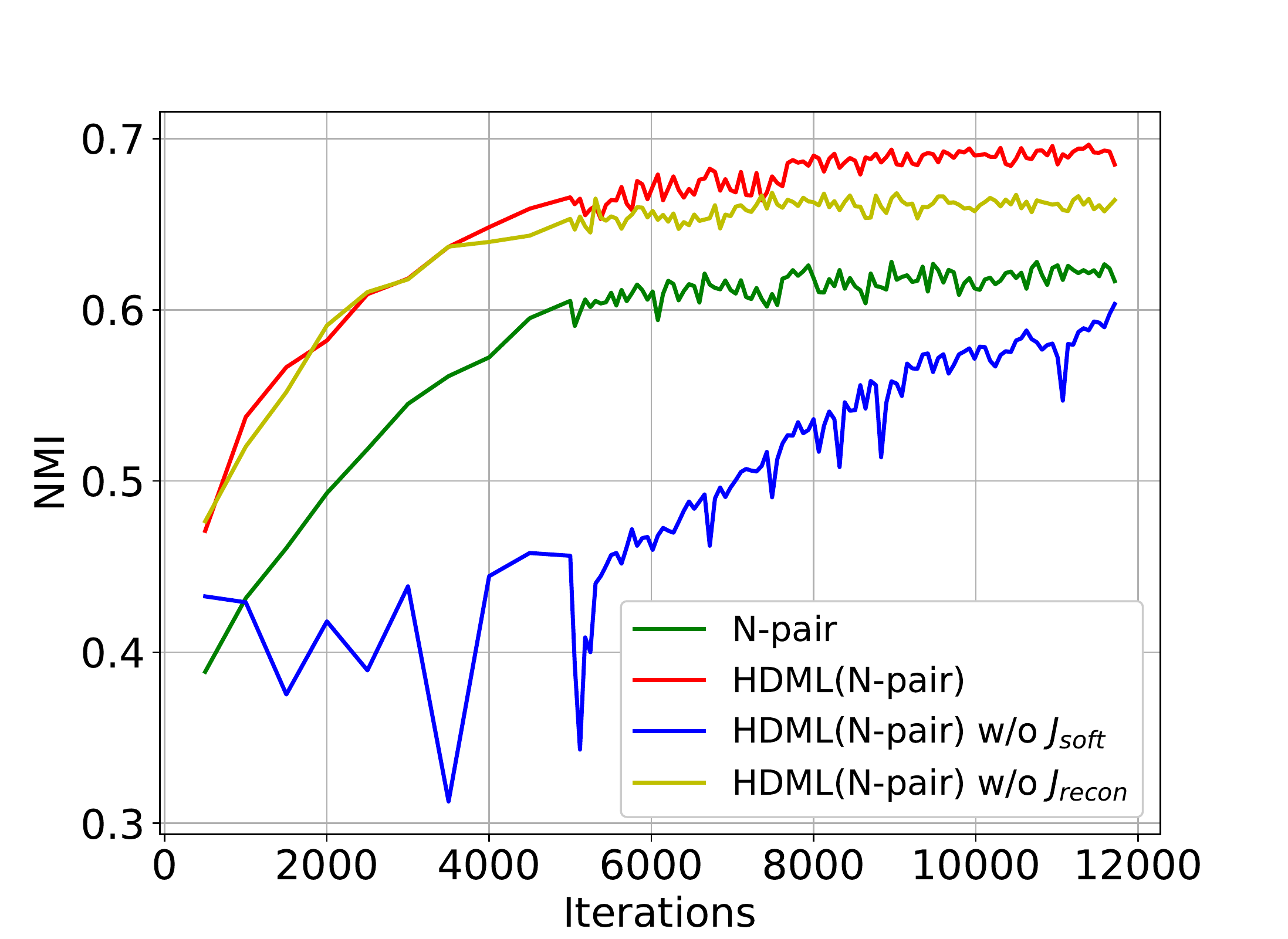}
\caption{Comparisons of different settings in the clustering task.}
\label{ablation_nmi}
\end{figure}

\begin{figure}[tb]
\centering
\includegraphics[width=\figwidth\textwidth]{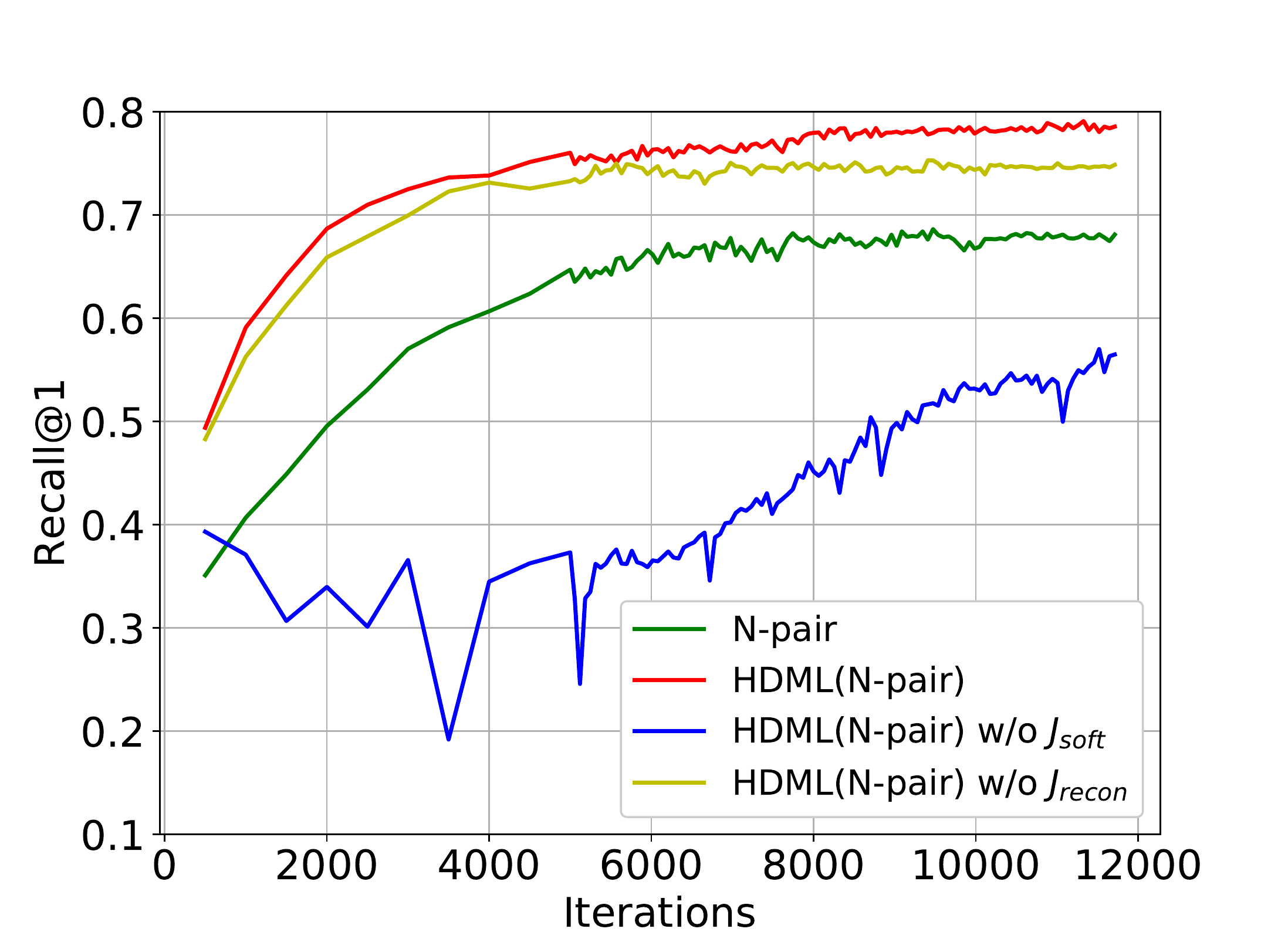}
\caption{Comparisons of different settings in the retrieval task.}
\label{ablation_recall}

\end{figure}

\begin{figure}[tb]
\centering
\includegraphics[width=\figwidth\textwidth]{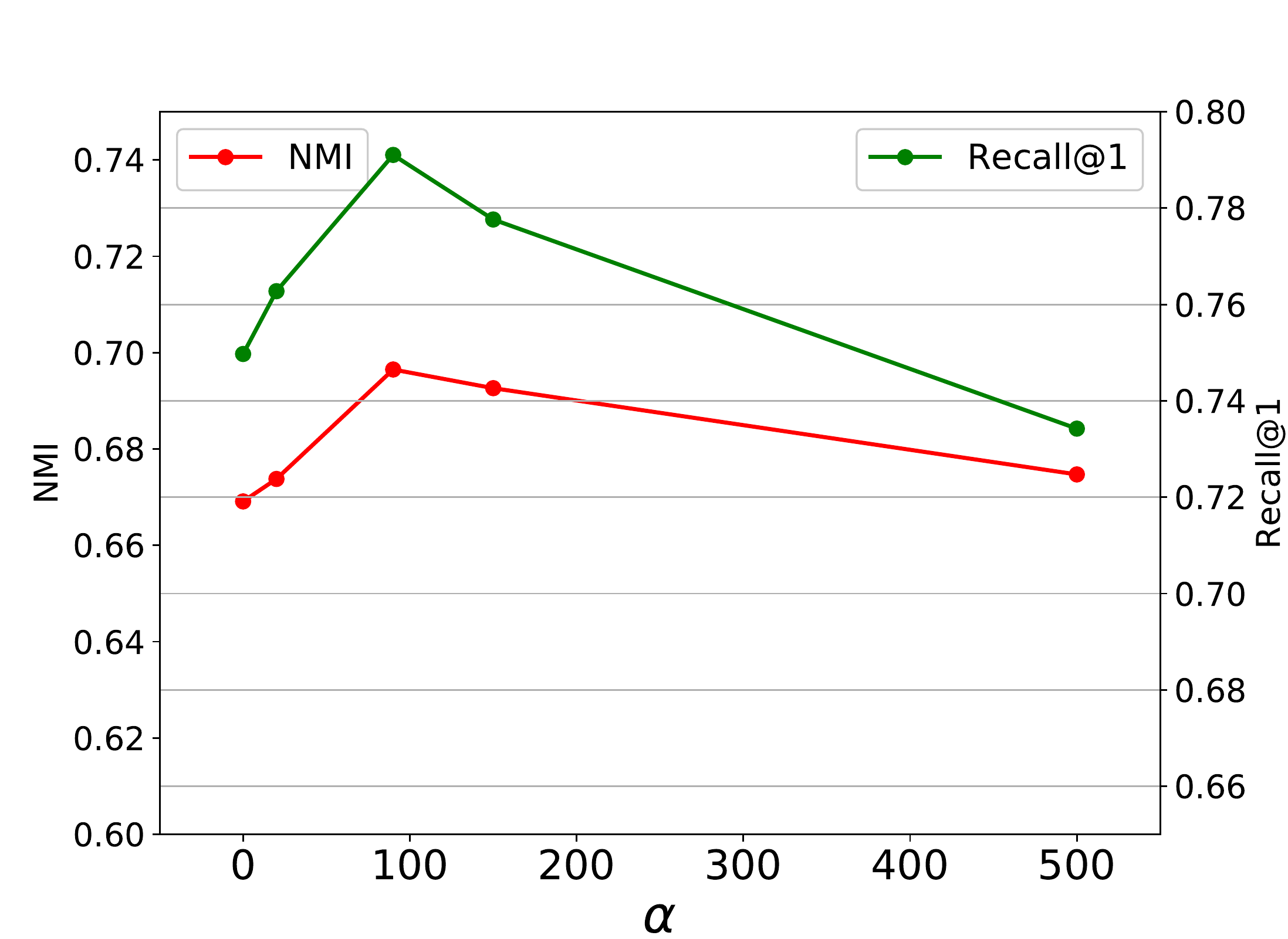}
\caption{Comparisons of converged results using different pulling factors in the clustering and retrieval task.}
\label{alpha}
\end{figure}

\begin{figure}[tb]
\centering
\includegraphics[width=\figwidth\textwidth]{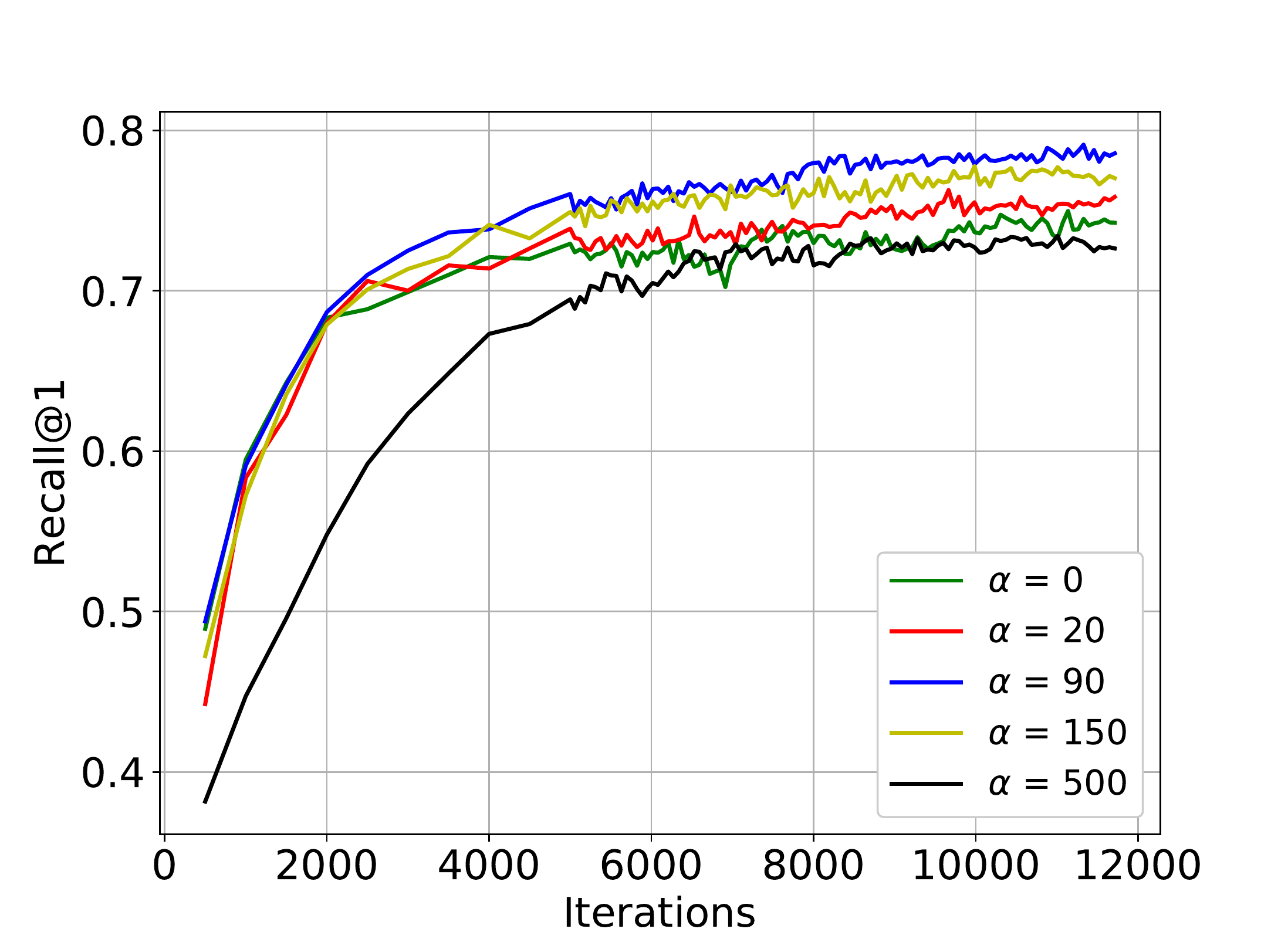}
\caption{Comparisons of using different pulling factors in the retrieval task.}
\label{alpha_recall}

\end{figure}

\begin{figure*}[tb]
\centering
\includegraphics[width=1\textwidth]{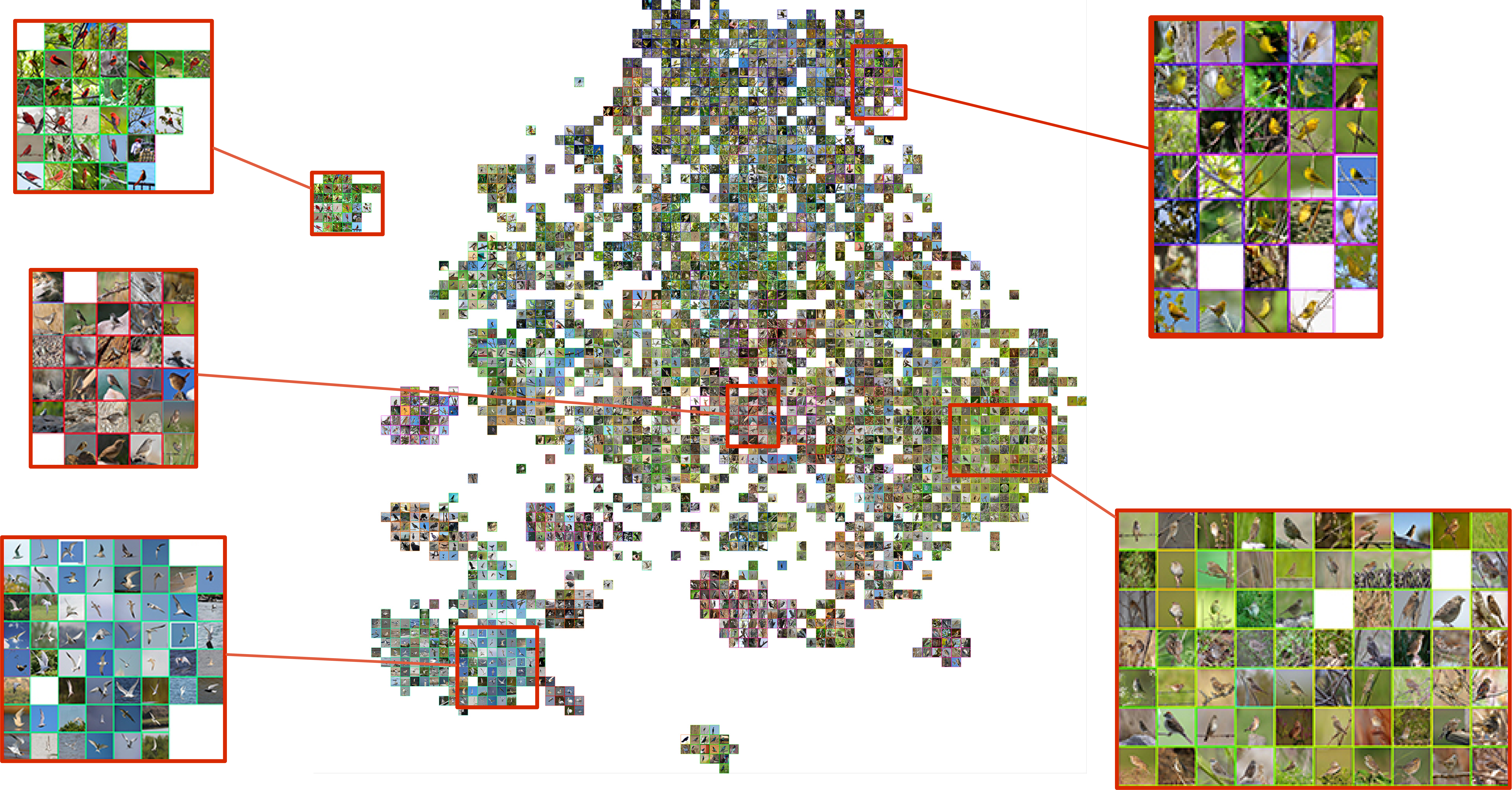}
\caption{Barnes-Hut t-SNE visualization~\cite{vandermaaten2014accelerating} of the proposed HDML (N-pair) method on the test split of CUB-200-2011, where we magnify several areas for a better view. The color of the boundary of each image represent the category. (Best viewed when zoomed in.)} 
\label{tsne_cub}
\end{figure*}

\textbf{Ablation Study:} We present the ablation study of the proposed method. We conducted all the following experiments on the Cars196 dataset with the N-pair loss, but we observe similar results with the triplet loss. 

Figures \ref{ablation_nmi} and \ref{ablation_recall} show the learning curves of different model settings in the clustering and retrieval task, including the baseline model, the proposed framework with the N-pair loss, the HDML framework without the softmax loss and the HDML framework without the reconstruction loss. We observe that the absence of the softmax loss results in dramatic performance reduction. This is because the synthetic samples might not preserve the label information, leading to inconsistent tuples. It is surprising that the proposed method without the reconstruction loss still achieves better results than the baseline. We speculate it is because the softmax layer itself learns to distinguish realistic synthetics from false ones in this situation.

Figures \ref{alpha} and \ref{alpha_recall} show the effect of different pulling factors. A larger $\alpha$ means we generate harder tuples each time, and $\alpha = 0$ means we do not apply hard synthesis at all.  We see that as $\alpha$ grows, the performance increases at first and achieves the best result at $\alpha = 90$, then gradually decreases. This justifies the synthesis of tuples with suitable and adaptive hardness. A too light hard synthesis may not fully exploit the underlying information, while a too strong hard synthesis may lead to inconsistent tuples and destroy the structure of the embedding space.

\begin{table}[tb] \small
\centering
\caption{Experimental results (\%) on the CUB-200-2011 dataset in comparison with other methods.}
\label{tab:cub}
\vspace{5pt}
\begin{tabular}{lccccccc}
\toprule
Method & NMI & $F_1$ & R@1 & R@2 & R@4 & R@8 \\
\midrule 
Contrastive  & 47.2 & 12.5    & 27.2 & 36.3 & 49.8 & 62.1   \\
DDML   & 47.3 & 13.1          & 31.2 & 41.6 & 54.7 & 67.1   \\
Lifted   & 56.4 & 22.6       & 46.9 & 59.8 & 71.2 & 81.5   \\
Angular     & 61.0 & 30.2       &  53.6 &  65.0 & 75.3 & 83.7   \\

 \midrule
Triplet  & 49.8 & 15.0        & 35.9 & 47.7 & 59.1 & 70.0   \\
Triplet hard& 53.4 & 17.9       & 40.6  & 52.3  & 64.2  & 75.0 \\  
DAML (Triplet) & 51.3 & 17.6 & 37.6 & 49.3 & 61.3 & 74.4 \\
HDML (Triplet)  & \textbf{55.1} & \textbf{21.9}    & \textbf{43.6} & \textbf{55.8} & \textbf{67.7} & \textbf{78.3} \\

\midrule
N-pair & 60.2 & 28.2     & 51.9 & 64.3 & 74.9 & 83.2 \\
DAML (N-pair) & 61.3 & 29.5 & 52.7 & 65.4 & 75.5 & 84.3 \\
HDML (N-pair)  & \color{red}\textbf{62.6} & \color{red}\textbf{31.6}    & \color{red}\textbf{53.7} & \color{red}\textbf{65.7} & \color{red}\textbf{76.7} & \color{red}\textbf{85.7} \\

\bottomrule
\end{tabular}
\end{table}

\begin{table}[tb] \small
\centering
\caption{Experimental results (\%) on the Cars196 dataset in comparison with other methods.}
\label{tab:cars}
\vspace{5pt}
\begin{tabular}{lccccccc}
\toprule
Method & NMI & $F_1$ & R@1 & R@2 & R@4 & R@8 \\
\midrule 
Contrastive     & 42.3 & 10.5       & 27.6 & 38.3 & 51.0 & 63.9    \\
DDML       & 41.7 & 10.9       & 32.7 & 43.9 & 56.5 & 68.8  \\
Lifted    & 57.8 & 25.1       & 59.9 & 70.4 & 79.6 & 87.0   \\

Angular     & 62.4 & 31.8       & 71.3 & 80.7 & 87.0 & 91.8   \\

 \midrule
Triplet       & 52.9 & 17.9         & 45.1 & 57.4 & 69.7 & 79.2   \\
Triplet hard    & 55.7 & 22.4      & 53.2  & 65.4  & 74.3  & 83.6  \\  
DAML (Triplet) & 56.5 & 22.9 & 60.6 & 72.5 & 82.5 & 89.9 \\
HDML (Triplet)  & \textbf{59.4} & \textbf{27.2}       & \textbf{61.0} & \textbf{72.6} & \textbf{80.7} & \textbf{88.5}\\
 \midrule
N-pair  & 62.7 & 31.8       & 68.9 & 78.9 & 85.8 & 90.9 \\
DAML (N-pair) & 66.0 & 36.4 & 75.1 & 83.8 & 89.7 & 93.5 \\
HDML (N-pair)    &\color{red}\textbf{69.7}  &\color{red}\textbf{41.6}      & \color{red}\textbf{79.1} & \color{red}\textbf{87.1} & \color{red}\textbf{92.1} & \color{red}\textbf{95.5} \\

\bottomrule
\end{tabular}
\end{table}

\begin{table}[tb] \small
\centering
\caption{Experimental results (\%) on the Stanford Online Products dataset in comparison with other methods.}
\label{tab:products}
\vspace{5pt}
\begin{tabular}{lcccccc}
\toprule
Method & NMI & $F_1$ & R@1 & R@10 & R@100 \\
\midrule 
Contrastive  & 82.4 & 10.1     & 37.5 & 53.9 & 71.0  \\
DDML    & 83.4 & 10.7      & 42.1 & 57.8 & 73.7  \\
Lifted      & 87.2 & 25.3       & 62.6 & 80.9 & 91.2\\
Angular      & 87.8 & 26.5       & 67.9 & 83.2 & 92.2 \\
 \midrule
Triplet     & 86.3 & 20.2    & 53.9 & 72.1 & 85.7\\
Triplet hard    & 86.7 & 22.1    & 57.8 & 75.3 & 88.1\\  
DAML (Triplet) & 87.1 & 22.3 & 58.1 & 75.0 & 88.0 \\
HDML (Triplet)   & \textbf{87.2} & \textbf{22.5}        &  \textbf{58.5} & \textbf{75.5} & \textbf{88.3}\\
 \midrule
N-pair   & 87.9 & 27.1        & 66.4 & 82.9 & 92.1\\
DAML (N-pair) & \color{red}89.4 & \color{red}32.4 & 68.4 & \color{red}83.5 & 92.3 \\
HDML (N-pair)  & \textbf{89.3}  &\textbf{32.2}       & \color{red}\textbf{68.7} & \textbf{83.2} & \color{red}\textbf{92.4}\\ 

\bottomrule
\end{tabular}
\end{table}

\textbf{Quantitative Results:}
We compared our model with several baseline methods, including the conventional contrastive loss~\cite{hadsell2006dimensionality} and triplet loss~\cite{weinberger2009distance}, more recent DDML~\cite{weinberger2009distance} and triplet loss with semi-hard negative mining~\cite{schroff2015facenet}, the state-of-the-art lifted structure~\cite{song2016deep}, N-pair loss~\cite{sohn2016improved} and angular loss~\cite{wang2017deep}, and the hard negative generation method DAML~\cite{duan2018deep}. We employed the proposed framework to the triplet loss and N-pair loss as illustrated before. We evaluated all the methods mentioned above using the same pre-trained CNN model for fair comparison.

Tables \ref{tab:cub}, \ref{tab:cars}, and \ref{tab:products} show the quantitative results on the CUB-200-2011, Cars196, and Stanford Online Products datasets respectively. Red numbers indicate the best results and bold numbers mean our method achieves better results than the associated method without HDML. We observe our proposed framework can achieve very competitive performance on all the three datasets in both tasks. Compared with the original triplet loss and N-pair loss, our framework can further boost their performance for a fairly large margin. This demonstrates the effectiveness of the proposed hardness-aware synthesis strategy. The performance improvement on the Stanford Online Products dataset is relatively small compared with the other two datasets. 
We think this difference comes from the size of the training set. 
Our proposed framework generates synthetic samples with suitable and adaptive hardness, which can exploit more information from a limited training set than conventional sampling strategies. This advantage becomes more significant on small-sized datasets like CUB-200-2011 and Cars196.

\textbf{Qualitative Results:}
Figure~\ref{tsne_cub} shows the Barnes-Hut t-SNE visualization~\cite{vandermaaten2014accelerating} of the learned embedding using the proposed HDML (N-pair) method. We magnify several areas for a better view, where the color on the boundary of each image represents the category. The test split of the CUB-200-2011 dataset contains 5,924 images of birds from 100 different species. The visual differences between two species tend to be very subtle, making it difficult for humans to distinguish. We observe that despite the subtle inter-class differences and large intra-class variations, such as illumination, backgrounds, viewpoints, and poses, our method can still be able to group similar species, which intuitively verify the effectiveness of the proposed HDML framework.

\section{Conclusion}
In this paper, we have presented a hardness-aware synthesis framework for deep metric learning. Our proposed HDML framework boosts the performance of original metric learning losses by adaptively generating  hardness-aware and label-preserving synthetics as complements to the training data. We have demonstrated the effectiveness of the proposed framework on three widely-used datasets in both clustering and retrieval task. In the future, it is interesting to apply our framework to the more general data augmentation problem, which can be utilized to improve a wide variety of machine learning approaches other than metric learning. 

\section*{Acknowledgement}
This work was supported in part by the National Natural Science Foundation of China under Grant 61672306,  Grant U1813218, Grant 61822603, Grant U1713214, and Grant 61572271.

{\small
\bibliographystyle{ieee_fullname}

\bibliography{egbib}
}

\end{document}